\documentclass[letterpaper, 10pt, conference]{ieeeconf}  %
\IEEEoverridecommandlockouts
\usepackage{times}
\usepackage{epsfig}
\usepackage{graphicx}
\usepackage{amsmath}
\usepackage{amssymb}
\usepackage{xcolor}
\usepackage{subcaption}
\usepackage{caption}
\usepackage{booktabs}
\usepackage[utf8]{inputenc}
\usepackage[nospace]{cite}
\usepackage[hidelinks]{hyperref}
\usepackage{enumerate}
\usepackage{flushend}
\usepackage{dsfont}
\usepackage{wrapfig}
\usepackage{cuted}

\usepackage[ruled,vlined,linesnumbered]{algorithm2e}

\graphicspath{{figures/}}

\newcommand{\new}[1]{\color{black} #1 \color{black}}

\newcommand{\old}[1]{}

\newcommand{\labelnotempty}[1]{
\def\temp{#1}\ifx\temp\empty
\else
    \label{#1}
\fi
}

\title{\LARGE \bf Monte-Carlo Tree Search for Efficient \\ Visually Guided Rearrangement Planning}

\author{
  {Yann Labbé}\,\textsuperscript{a,b},
    {Sergey Zagoruyko}\,\textsuperscript{a,b},
    {Igor Kalevatykh}\,\textsuperscript{a,b},
    {Ivan Laptev}\,\textsuperscript{a,b}, \\
    {Justin Carpentier}\,\textsuperscript{a,b},
    {Mathieu Aubry}\,\textsuperscript{c} and
    {Josef Sivic}\,\textsuperscript{a,b,d}
\thanks{This work was partially supported by the DGA
RAPID projects DRAAF and TABASCO, the MSR-Inria
joint lab, the Louis Vuitton - ENS Chair on Artificial
Intelligence, the HPC resources from GENCI-IDRIS (Grant 011011181), the ERC grant LEAP (No. 336845), the CIFAR
Learning in Machines\&Brains program, the European
Regional Development Fund under the project IMPACT (reg.
no. CZ.02.1.01/0.0/0.0/15 003/0000468) and the French government under management of Agence Nationale de la Recherche as part of the "Investissements d'avenir" program, reference ANR-19-P3IA-0001 (PRAIRIE 3IA Institute).}
\thanks{\textsuperscript{a}\,D\'epartement d’informatique de l’ENS, \'Ecole normale sup\'erieure, CNRS, PSL Research University, 75005
  Paris, France, \textsuperscript{b}\,INRIA, France,%
  \textsuperscript{c}\,LIGM (UMR 8049), \'Ecole des Ponts, UPE, France,%
  \textsuperscript{d}\,Czech Institute of Informatics, Robotics and Cybernetics, Czech Technical University in Prague.%
}%
\thanks{Corresponding author: yann.labbe@inria.fr}
}

\begin{document}

\maketitle

\begin{abstract}
We address the problem of visually guided rearrangement planning with many movable objects, i.e., finding a sequence of actions to move a set of objects from an initial arrangement to a desired one, while relying on visual inputs coming from an RGB camera.
To do so, we introduce a complete pipeline relying on two key contributions. %
First, we introduce an efficient and scalable rearrangement planning method, based on a Monte-Carlo Tree Search exploration strategy. %
We demonstrate that because of its good trade-off between exploration and exploitation our method (i) scales well with the number of objects while (ii) finding solutions which require a smaller number of moves compared to the other state-of-the-art approaches. Note that on the contrary to many approaches, we do not require any buffer space to be available.
Second, to precisely localize movable objects in the scene, we develop an integrated approach for robust multi-object workspace state estimation from a single uncalibrated RGB camera using a deep neural network trained only with synthetic data. 
We validate our multi-object visually guided manipulation pipeline with several experiments on a real UR-5 robotic arm by solving various rearrangement planning instances, requiring only 60\,ms to compute the plan to rearrange 25 objects. %
In addition, we show that our system is insensitive to camera movements and can successfully recover from external perturbations. Supplementary video, source code and pre-trained models are available at \href{https://ylabbe.github.io/rearrangement-planning}{https://ylabbe.github.io/rearrangement-planning}.

\end{abstract}

\section{Introduction}
\label{sec:introduction}
\begin{figure}
  \centering
  \includegraphics[width=\columnwidth]{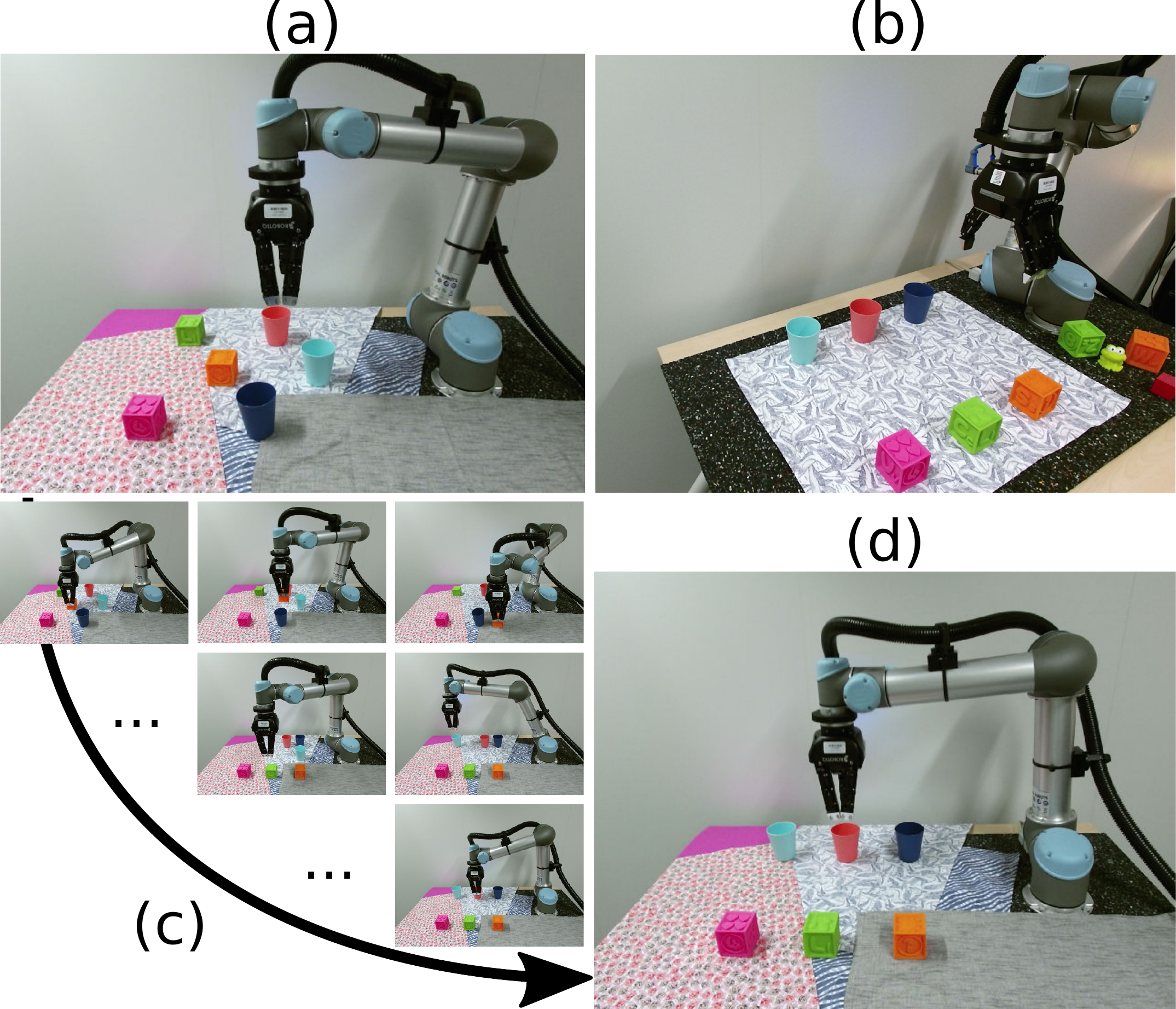}
  \caption{\small {\bf Visually guided rearrangement planning. } Given a source (a) and target (b)
            RGB images depicting a robot and multiple movable objects, 
            our approach estimates the positions of objects in the scene without the need for explicit camera calibration and efficiently finds a sequence of
            robot actions (c) to re-arrange the scene into the target
            scene. Final object configuration after re-arrangement by the robot is shown in (d).
             }
  \label{fig:teaser}
  \vspace{-2.0em}
\end{figure}

Using
a robot to clean up a room is a dream shared far beyond the robotics community.
This would require a robot to both localize and re-arrange many objects.
Other industrial scenarios, such as sorting and packing objects on a production line or car assembly tasks, share similar objectives and properties. 
This paper presents an integrated approach that makes a step towards the efficiency, scalability and robustness required for solving such rearrangement planning tasks. 
Fig.\,\ref{fig:teaser} shows an example of the problem we consider, where objects have to be moved from an initial position to a target one. The current and target states are described only by a single image taken from an uncalibrated RGB camera.

Rearrangement planning has a long history in robotics\,\cite{alami1990geometrical,simeon2004manipulation,lavalle2006planning,stilman2007manipulation,latombe2012robot,kaelbling2011hierarchical} and remains an active research topic\,\cite{havur2014geometric,krontiris2015dealing,dantam2016incremental,Han2017-sa} in the motion planing community.
The goal is to find a sequence of \textit{transit} and \textit{transfer} motions\,\cite{alami1990geometrical,latombe2012robot} to move a set of objects from an initial arrangement to a target arrangement, while avoiding collisions with the environment. 
This leads to a complex sequential decision process, whose complexity depends on the number of objects to move, on the free-space available around the objects, and the robot kinematics.

Several solutions have been proposed in the literature which can be roughly classified into two groups.
Methods in the first group~\cite{kaelbling2011hierarchical,dantam2016incremental,krontiris2015dealing,Srivastava2014-sc,Garrett2018-te,Han2017-sa} 
rely on the \textit{task} and \textit{motion} planning hierarchy where a
high-level task planner is combined with a local motion
planner\cite{lavalle2006planning}. Methods in the second group~\cite{simeon2004manipulation,stilman2007manipulation,krontiris2016efficiently,Mirabel2016-zu} aim at solving a single unified formulation of the problem by using classic sample-based algorithms such as Probabilistic RoadMap (PRM) or Rapidly-Exploring Random Tree (RRT)\,\cite{lavalle2006planning} or use advanced optimization strategies to solve a unique optimization instance~\cite{toussaint2015logic}.

While methods from both groups have been shown to work well in practice with few objects, existing methods do not scale to a large set of objects, 
because the number of possible action sequences increases exponentially with the number of objects to move.
{Some recent methods \cite{krontiris2015dealing,krontiris2016efficiently,Han2017-sa} scale better with the number of objects but these methods either only focus on feasibility, producing solutions with sub-optimal number of grasps~\cite{krontiris2015dealing}, or are limited to specific constrained scenarios, for example, with explicitly available buffer space~\cite{Han2017-sa} or strict constraints of monotony (i.e. an object can be moved only once during the plan).
}

{In this work we describe an efficient and generic approach for rearrangement planning that overcomes these limitations: (i) it scales well with the number of objects, by taking only 60\,ms to plan complex rearrangement scenarios for multiple objects, and (ii) it can be applied to the most challenging table-top re-arrangement scenarios, not requiring explicit buffer space.}
Our approach is based on Monte-Carlo Tree Search~\cite{munos2014bandits}, which allows us to lower the combinatorial complexity of the decision process and to find an optimistic number of steps to solve the problem, by making a compromise between exploration (using random sampling) and exploitation (biasing the search towards the promising already sampled action sequences to cut off some of the search directions). %

\begin{figure}
  \centering
  \includegraphics[width=0.8\columnwidth]{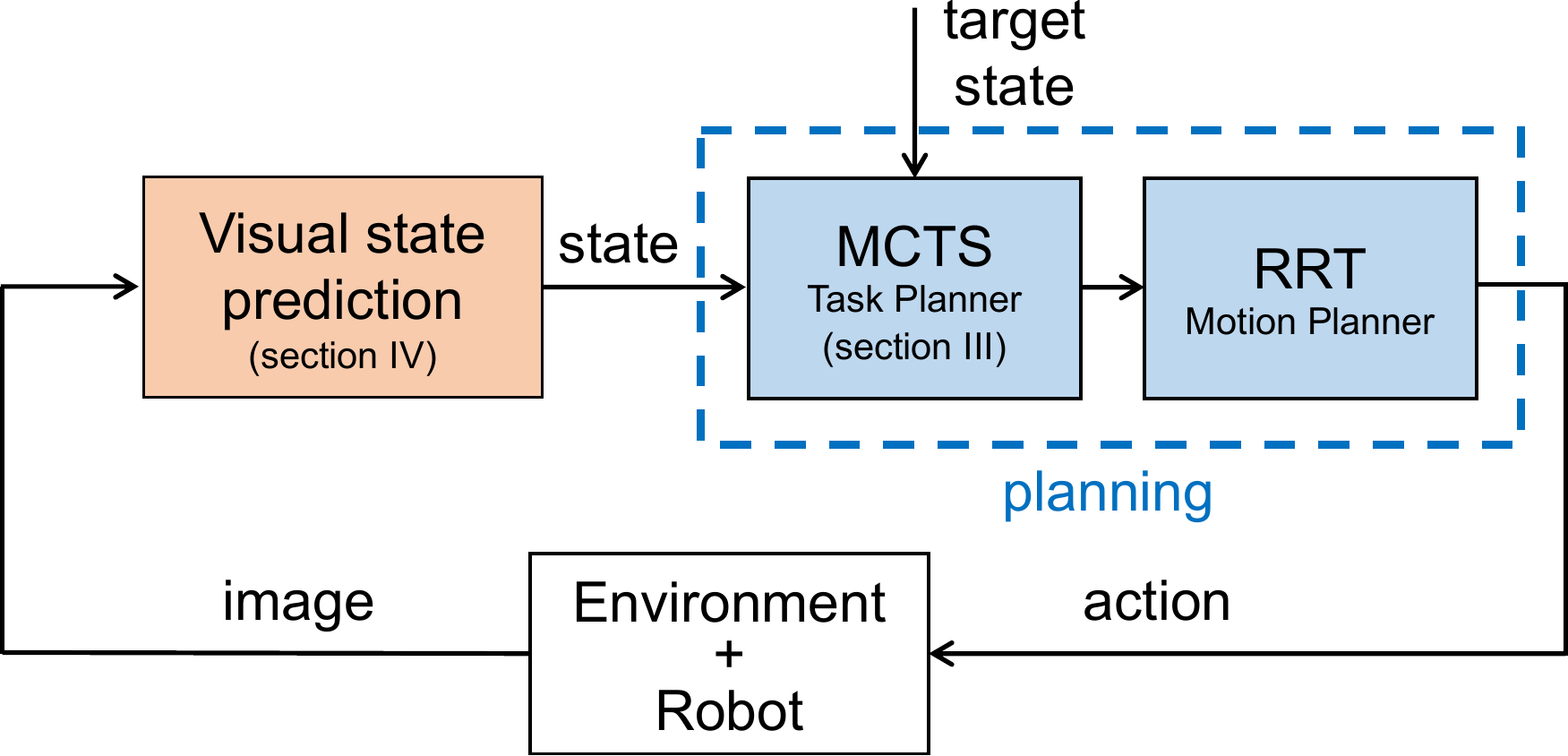}
  \caption{\small {\bf Approach overview.} 
    Given an image of the scene, the visual state prediction module outputs a list of objects and their coordinates in the robot coordinate frame.
    Together with a target state, these serve as inputs to the task and motion planning module which combines Monte-Carlo Tree Search with a standard robot motion planning algorithm.}
  \label{fig:pipeline}
  \vspace{-2.0em}
\end{figure}

To demonstrate the benefits of our planning approach in real scenarios we also introduced a multi-object calibration-free deep neural network architecture for object position estimation. It is trained entirely from synthetic images and, {compared to other currently available methods~\cite{Garrido-Jurado2014-uh,Li2018-fp,Xiang2018PoseCNNAC,Tremblay2018DeepOP}, does not use markers or require known CAD models of the specific observed object instances.} To the best of our knowledge, the approach we present is the first one able to locate multiple objects in such difficult and generic conditions. This is of high practical interest, since it allows to perform the full task using only a single non-calibrated, or even hand-held, RGB camera looking at the robot.

Our complete pipeline for visually guided rearrangement planning, illustrated in Fig.~\ref{fig:pipeline}, is composed of three main stages.
The goal of the first stage, visual state prediction (section~\ref{sec:vision}), is to estimate the positions of multiple objects relative to a robot given a single non-calibrated image of the scene. 
The second stage (Sec.~\ref{sec:mcts}), is our MCTS-based task planner: at each planning step, this module chooses which object to move and computes its new desired position in the workspace. The last stage is a standard RRT-based local motion planner which plans robot movements given the high-level plan computed by the MCTS planner.

\section{Related work}
\label{sec:related_work}

We build our framework on results in robotics, search algorithms and computer vision, which we review below.

\noindent \textbf{Rearrangement planning} is \mbox{NP-hard}~\cite{wilfong1991motion}. %
{As a result, standard hierarchical~\cite{kaelbling2011hierarchical,dantam2016incremental,krontiris2015dealing,Garrett2018-te} and randomized methods~\cite{simeon2004manipulation,stilman2007manipulation,Mirabel2016-zu} for solving 
general manipulation planning problems do not scale well with the number of objects.
The most efficient and scalable high-level planners only address specific constrained set-ups leveraging the structure
of the rearrangement problem \cite{krontiris2015dealing,krontiris2016efficiently,Han2017-sa}. In addition, they often 
focus on feasibility but do not attempt to
find high-quality solutions with a low number of object moves~\cite{krontiris2015dealing,krontiris2016efficiently}. 
For instance, some methods~\cite{krontiris2016efficiently} only 
consider the monotone problem instances, where each object can be grasped at most once.
In contrast, our method finds high-quality plans but also addresses the more general cases of non-monotone re-arrangement problems, which are known to be significantly harder~\cite{stilman2007manipulation}. 
Others works have looked at finding optimal plans~\cite{Han2017-sa} but address only constrained set-ups that have available buffer space (i.e. space that does not overlap with the union of the initial and target configurations), noting that solving the general case without available buffer space is significantly harder~\cite{Han2017-sa}.
In this work, we address this more general case and describe an approach that efficiently finds high-quality re-arrangement solutions without requiring any available buffer space.
In addition and unlike previous works~\cite{krontiris2015dealing,krontiris2016efficiently,Han2017-sa}, we also propose a complete system able to operate from real images in closed loop.}

\textbf{Search algorithms.}
The problem of rearrangement planning can be posed as a tree search.
Blind tree search algorithms such as Breadth-First search (BFS) \cite{russel2003} can be used to iteratively expand nodes of a tree until a goal node is found, but these methods 
do not exploit information about the problem (e.g. a cost or reward) to select which nodes to expand first,
and typically scale exponentially with the tree depth.
Algorithms such as greedy BFS \cite{russel2003} allow to exploit a reward function to drive the exploration of
the tree directly towards nodes with high reward,
but might get stuck in a local mimima.
Other algorithms such as $A^\star$ \cite{astar} can better estimate the promising branches using additional hand-crafted heuristic evaluation function.
We choose Monte-Carlo Tree Search over others, because it only relies on a reward and 
iteratively learns a heuristic (the value function) which allows to efficiently balance between
exploration and exploitation.
It has been used in related areas to solve planning and routing for ground transportation~\cite{paxton2017combining} and to guide the tree-search in cooperative manipulation~\cite{toussaint2017multi}.
MCTS is also at the core of AlphaGo, the first system able to achieve human performance in the game of Go~\cite{Silver2016MasteringTG}, where it was combined with neural networks to speed-up the search.
These works directly address problems whose action space is discrete by nature. In contrast, the space of possible object arrangements is infinite. We propose a novel discrete action parameterization of the rearrangement planning problem which allows us to efficiently apply MCTS.

\noindent \textbf{Vision-based object localization}.
In robotics, fiducial markers are commonly used for detecting the objects and
predicting their pose relative to the camera \cite{Garrido-Jurado2014-uh} but their use limits the type of
environments the robot can operate in.
This constraint can be removed by using a trainable object detector architecture
\cite{wu2018pose, Xiang2018PoseCNNAC,Tremblay2018DeepOP,
  zeng2016multi, Hodan2016-lz,Li2018-fp}. However, these
methods often require gathering training data for the target objects at hand, which is often time consuming and requires the knowledge of the object (e.g. in the
form its 3D model) beforehand.
In addition, these methods estimate the pose of the objects in the frame of the
camera and using these predictions for robotic manipulation requires calibration
of the camera system with respect to the robot. The calibration procedure
\cite{Horaud1995-fl, Heller2011-so} is time-consuming and must be
redone each time the camera is moved. More recently, \cite{Loing2018} proposed
to directly predict the position of a single object in the robot coordinate frame by training a deep network on hundreds of thousands of synthetically generated images
using domain randomization \cite{Tobin2017DomainRF, Loing2018, Sadeghi2017-dr, James2018-db}. We build on the work~\cite{Loing2018} and extend it for predicting the 2D positions of {\em multiple objects with unknown dimensions} relative to the robot.

\section{Visual scene state prediction with multiple objects}
\label{sec:vision}

In this section, we detail the visual state prediction stage.
Our visual system takes as input a single photograph of a scene taken from an uncalibrated
camera and predicts a workspace state that can then be used for rearrangement-planning. More precisely it outputs the 2D positions of a variable number of objects expressed in the coordinate system of  the robot. 
This problem is difficult because the scene can contain a variable number
of objects, placed on different backgrounds, in variable illumination conditions, and observed from
different viewpoints, as illustrated in Fig.~\ref{fig:teaser}.
In contrast to~\cite{Tobin2017DomainRF}, we do not assume that the different types of objects are known at training time.
In contrast to state-of-the-art pose estimation techniques in RGB images
  \cite{Garrido-Jurado2014-uh,Li2018-fp}, we do not use markers and do
  not assume the CAD models of the objects are known at test time.

\begin{figure*}
  \centering
  \includegraphics[width=\textwidth]{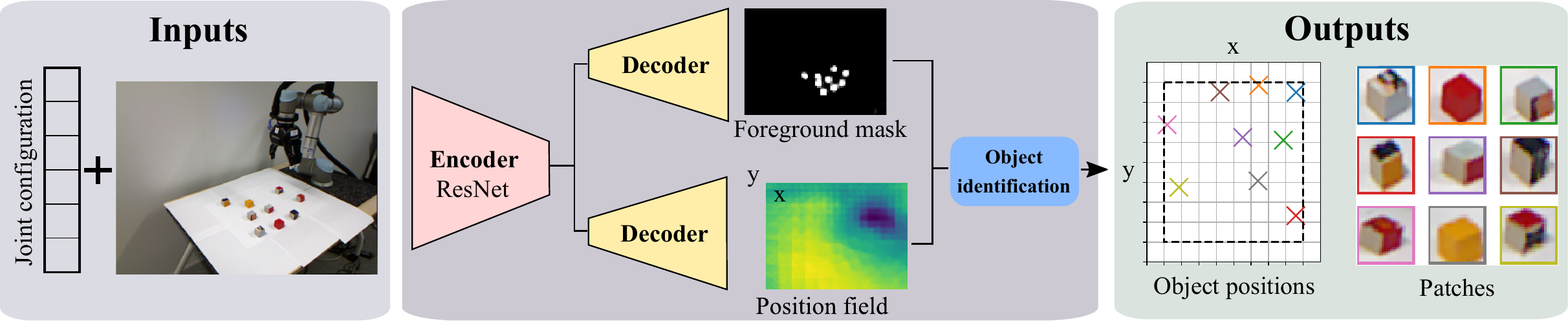}
  \caption{\small {\bf The visual recognition system.} The input is an image of the scene captured by an uncalibrated camera together with the joint configuration vector of the depicted robot.  Given this input a convolutional neural network (CNN) predicts the foreground-background object segmentation mask and a dense position field that maps each pixel of the (downsampled) input image to a 2D coordinate in a frame centered on the robot. The estimated masks and position fields are then combined by the object identification module to identify individual object instances. The output is a set of image patches associated with the 2D position of the object in the workspace.}
    
  \label{fig:visual_system}
  \vspace{-1em}
\end{figure*}

To address these challenges, we design a visual recognition system that does not
require explicit camera calibration and outputs accurate 2D positions. 
Moreover, even if we deploy our system on a real robot, we show that it can be trained entirely from synthetic data using domain randomization~\cite{Tobin2017DomainRF}, avoiding the need for real training images.
Also, our system does not require any tedious camera calibration because it is trained to predict positions of objects directly in the coordinate frame of the robot, effectively using the robot itself, which is visible in the image, as an (known) implicit calibration object. This feature is important for applications in unstructured environments such as construction sites containing multiple unknown objects and moving cameras for instance.
Our recognition system is summarized in Fig.~\ref{fig:visual_system} and in the rest of this section, we present in more details the different components. %

\subsection{Position prediction network}
\label{sec:network}

In this section, we give details of the network for
predicting a dense 2D position field and an object segmentation mask. The 2D position field maps each input pixel to a 2D coordinate frame of the robot acting as implicit calibration.

\noindent \textbf{Architecture.}
Our architecture is based on ResNet-34 \cite{He2016DeepRL}. We remove the average
pooling and fully connected layers and replace them by two independent decoders.
Both decoders use the same architecture: four transposed convolution layers with
batch normalization and leaky ReLU activations in all but the last layer.
The resolution of the input image is 320 $\times$ 240 and the spatial
resolution of the output of each head is 85 $\times$ 69.
We add the 6D joint configuration vector of the robot as input to the network by copying it into a tensor of size 
320 $\times$ 240 $\times$ 6, and simply concatenating it with the three channels of the input image.
The two heads predict an object mask and a 2D position field which are visualized in Fig.~\ref{fig:visual_system}. In addition, we found that predicting depth and semantic segmentation during training increased the localization accuracy at test time. These modalities are predicted using two additionnal decoders with the same architecture.

\begin{figure}
  \centering
  \includegraphics[width=\linewidth]{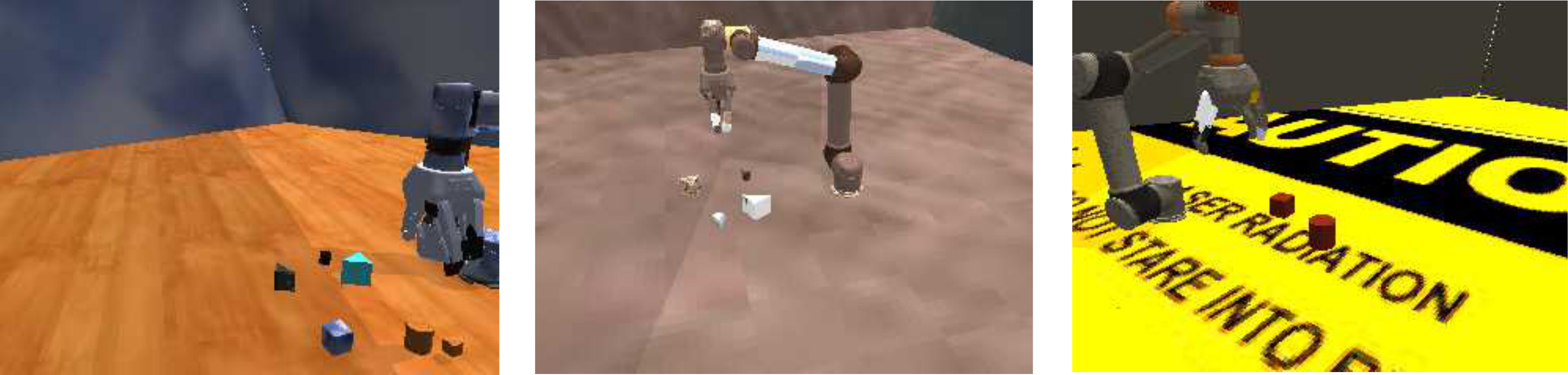}
  \caption{\small Examples of synthetic training images. We generate images
    displaying the robot and a variable number of objects in its workspace. The images are 
    taken by cameras with different viewpoints and depicting large scene appearance variations.}
  \label{fig:train_images}
  \vspace{-2em}
\end{figure}
\noindent \textbf{Synthetic training data.}
Following~\cite{Loing2018, Tobin2017DomainRF, Sadeghi2017-dr, James2018-db}, we use domain randomization for
training our network without requiring any real data. We generate two
million images displaying the robot and a variable number of objects with various shapes\,(cubes, cylinders and triangles) in its
workspace. In each scene, we randomly sample from 1 up to 12 objects, with various dimensions  between 2.5 and 8\,cm. Examples of training images are shown in Fig.~\ref{fig:train_images}. 
Randomization parameters include the textures of the robot and objects, the position
of the gripper, the position, orientation and field of view of the camera,
the positions and intensities of the light sources and their
diffuse/ambient/specular coefficients. %

\noindent \textbf{Training procedure.}
We train our network by minimizing the following loss:
$\mathcal{L} = \mathcal{L}_{\mathrm{pos}} + \mathcal{L}_{\mathrm{mask}} + \mathcal{L}_{\mathrm{segm}} + \mathcal{L}_{\mathrm{depth}},$
where the individual terms are explained next. For the position field loss we use \mbox{$\mathcal{L}_{\mathrm{pos}}=\sum_{i, j} \delta_{i, j}
\left[ (\hat{x}_{i,j} - x_{i,j})^2 + (\hat{y}_{i,j} - y_{i,j})^2 \right]$} where $(i, j)$ are the pixel coordinates in the output;
$\delta_{i,j}$ is the binary object mask; %
 ${x_{i,
  j}}$, $y_{i, j}$ are the ground truth 2D coordinates of the center of the object (that appears
at pixel $(i,j)$) and $\hat{x}_{i, j},
\hat{y}_{i,j}$ are the components of the predicted position field. %
For $\mathcal{L}_{\mathrm{mask}}$, $\mathcal{L}_{\mathrm{segm}}$ and
$\mathcal{L}_{\mathrm{depth}}$ we respectively use the following standard losses:
binary cross entropy loss, cross entropy loss and mean squared error. These
  losses are computed pixel-wise. Note that depth is not used to
  estimate object positions at test time. $L_{\mathrm{segm}}$ and $L_{\mathrm{depth}}$ are auxiliary
  losses used only for training, similar to \cite{James2018-db}.
We use the Adam optimizer~\cite{Kingma2015AdamAM} and train the network for 20
epochs, starting with a learning rate of $10^{-3}$ and decreasing it to $10^{-4}$
after 10 epochs.

\subsection{Identifying individual objects}
The model described above predicts a dense 2D position field and an object mask but does not distinguish individual objects in the scene.
Hence, we use the following procedure to group pixels belonging to each individual object.
Applying a threshold to the predicted mask yields a binary object segmentation. The corresponding pixels of the 2D position field provide a point set in the
robot coordinate frame. We use the mean-shift algorithm \cite{Comaniciu2002-ux} to cluster the 2D points corresponding to the different objects and obtain an estimate of the position of each object.
The resulting clusters then identify pixels belonging to each individual object providing instance segmentation of the input image.
We use the resulting instance segmentation to extract patches that describe the appearance of each object in the scene.

\subsection{Source-Target matching}
  \label{sec:matching}
  To perform rearrangement, we need to associate each object in the current image to an object in the target configuration. To do so, we use the extracted image patches. We designed a simple procedure to obtain matches robust to the exact position of the object within the patches, their background and some amount of viewpoint variations. We rescale patches to 64$\times$64 pixels and extract conv3 features of an AlexNet network trained for ImageNet classification. We finally run the Hungarian algorithm to find the one-to-one matching between source and target patches maximizing the sum of cosine similarities between the extracted features. \new{We have tried using features from different layers, or from the more recent network ResNet. We found the conv3 features of AlexNet to be the most suited for the task, based on a qualitative evaluation of matches in images coming from the dataset presented in Sec.~\ref{sec:exvis}}. Note that our patch matching strategy assumes all objects are visible in the source and target images.

\vspace{0.3cm}
\section{Rearrangement Planning with Monte-Carlo Tree Search (MCTS)}
\label{sec:mcts}

Given the current and target arrangements, the MCTS task planner has to compute a sequence of pick-and-place actions that transform the current arrangement into the target one. In this section, we first review Monte-Carlo Tree Search and then explain how we adapt it for rearrangement planning.

\subsection{Review of Monte-Carlo Tree Search}
\label{sec:reviewMCTS}
The MCTS decision process is represented by a tree, where each node is associated to a state $s$, and each edge represents a discrete action $a=\{1,...,N\}$.
For each node in the tree, a reward function $r(s)$ gives a scalar representing the level of accomplishment of the task to solve. Each node stores the number of times it has been visited $n(s)$ and a cumulative reward $w(s)$. The goal of MCTS is to find the most optimistic path, i.e.~the path that maximizes the expected reward, starting from the root node and leading to a leaf node solving the task. MCTS is an iterative algorithm where each iteration is composed of three stages. 

During the \textit{selection} stage, an action is selected using the Upper Confidence Bound\,(UCB) formula:
\begin{equation}
  U(s,a) = Q(s,a)+ c \sqrt{\frac{2\log n(s)}{n(f(s,a))}} ,
  \label{eq:uct_uniform}
\end{equation}

\noindent where $f(s,a)$ is the child node of $s$ corresponding to the edge $(s,a)$ and $Q(s,a)=\frac{w(f(s,a))}{n(f(s,a))}$ is the expected value at state $s$ when choosing action $a$. 
The parameter $c$ controls the trade-off between exploiting states with high expected value (first term in~\eqref{eq:uct_uniform}) and exploring states with low visit count (second term in~\eqref{eq:uct_uniform}). We found this trade-off is crucial for finding good solutions in a limited number of iterations as we show in Sec.~\ref{sec:exmcts}. The optimistic action selected $a_\text{selected}$ is the action that maximizes $U(s,a)$ given by~\eqref{eq:uct_uniform}. Starting from the root node, this stage is repeated until an expandable node (a node that has unvisited children) is visited. Then, a random unvisited child node $s^\prime$ is added to the tree in the \textit{expansion stage}. The reward signal $r(s^\prime)$ is then back-propagated towards the root node in the \textit{back-propagation stage}, where the cumulative rewards $w$ and visit counts $n$ of all the parent nodes are updated. The search process is run iteratively until the problem is solved or a maximal number of iterations is reached.

\subsection{Monte-Carlo Tree Search task planner}
\label{sec:mcts_task_planner}

\begin{figure}
  \centering
  \includegraphics[width=1.0\columnwidth]{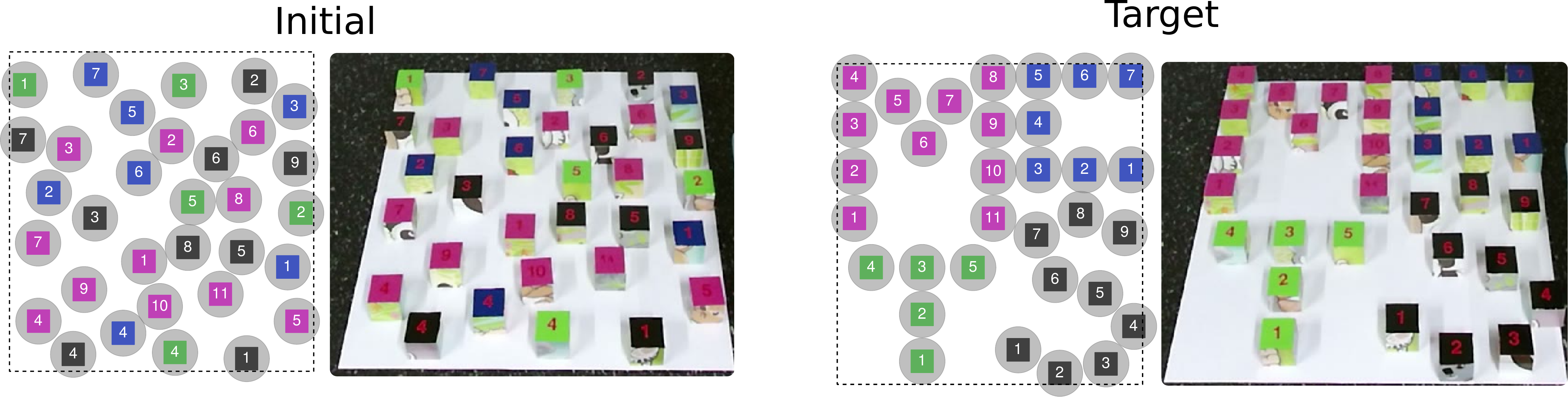}
  \caption{\small Examples of source and target object configurations.
    The planner has to find a sequence of actions (which object to move and where to displace it inside the workspace), while avoiding collisions with other objects.
    Workspace limits are shown as dashed lines. Grey circles depict the
    collision radius for each object. We demonstrate our method solving this
    complex rearrangement problem in the {\bf supplementary video.} Here source and target states are known (state estimation network is not used).
  }
  \label{fig:configs}
\end{figure}

We limit the scope to tabletop rearrangement planning problems with overhand grasps but our solution may be applied to other contexts.
We assume that the complete state of any object is given by its 2D position in the workspace, and this information is sufficient to grasp it. The movements are constrained by the limited workspace, and actions should not result in collisions between objects. The planner has to compute a sequence of actions which transform the source arrangement into the target arrangement while satisfying these constraints. Each object can be moved multiple times and we do not assume that explicit buffer space is available.
An example of source and target configurations is depicted in Fig.~\ref{fig:configs}. We now detail the action parametrization and the reward.

Let $\{C_i\}_{i=1,..,N}$ denote the list of 2D positions that define
the current arrangement with $N$ objects,
$\{I_i\}_{i}$ and $\{T_i\}_i$ the initial and target arrangements respectively, which are fixed for a given rearrangement problem. MCTS state corresponds to an arrangement $s = \{C_i\}_{i}$.

As a reward $r(s)$ we use the number of objects located within a small distance of their target position:
\begin{equation}
r(s) = \sum_{i=1}^N R_i \text{ with } R_i = \begin{cases} 1& \text{if } ||C_i - T_i||_2 \leq \epsilon \\ 0& \text{otherwise.}\end{cases}, \label{eq:reward}
\end{equation}
where $N$ is the number of objects, $C_i$ is the current location of object $i$, $T_i$ is the target location of object $i$ and $\epsilon$ is a small constant.

We define a discrete action space with $N$ actions where each action corresponds to one pick-and-place motion moving one of the $N$ objects. The action is hence pararametrized by 2D picking and placing positions defined by the function \textsc{GET\_MOTION} outlined in detail in Algo.~\ref{algo:action_param}. The input to that function is the current state $\{C_i\}_i$, target state $\{T_i\}_i$ and the chosen object $k$ that should be moved. The function proceeds as follows (please see also Algo.~\ref{algo:action_param}). First, if possible, the object $k$ is moved directly to it's target $T_k$ (lines~\ref{start:is_move_valid}-\ref{end:is_move_valid} in Algo.~\ref{algo:action_param}), otherwise the obstructing object $j$ which is the closest to $T_k$ (line~\ref{object_j} in Algo.~\ref{algo:action_param}) is moved to a position inside the workspace that does not overlap with $T_k$ (lines~\ref{start:move_j}-\ref{end:move_j} in Algo.~\ref{algo:action_param}). The position $P$ where to place $j$ is found using random sampling (line~\ref{start:move_j}). For collision checks, we consider simple object collision models with fixed radiuses as depicted in Figure~\ref{fig:configs}. If no suitable position is found, no objects are moved (line~\ref{move_k}). Note that additional heuristics could be added to the action parametrization to further improve the quality of the resulting solutions and to reduce the number of iterations required. Examples include (i) avoiding to place $j$ at target positions of other objects and (ii) restricting the search for position $P$ in a neighborhood of $C_j$. The parameters of the pick-and-place motion for a given state and MCTS action are computed only once during the expansion stage and then cached in the tree and recovered once a solution is found.

\begin{algorithm}
  \footnotesize
  \caption{Action Parametrization} \label{algo:action_param}
  \SetKwInOut{Input}{input}\SetKwInOut{Output}{output}
  \SetKwFunction{GETPP}{GET\_MOTION}
  \SetKwFunction{CollisionCheck}{IS\_MOVE\_VALID}
  \SetKwFunction{FindSpace}{FIND\_POSITION}
  \SetKwFunction{FindClosest}{FIND\_CLOSEST\_OBJECT}
  \SetKwProg{Fn}{function}{\string:}{}

  \Fn{\GETPP{$\{C_i\}_{i}$, $\{T_i\}_{i}$, $k$}}{

    /* Check if object $k$ can be moved to $T_k$ */

  \eIf{\CollisionCheck($\{C_i\}_{i\neq k}$, $C_k$, $T_k$)}{ \label{start:is_move_valid}
    \KwRet ($C_k$, $T_k$) \label{end:is_move_valid}
  }{

    /* Move obstructing object $j$ to position $P$*/

    $j$ = \FindClosest($\{C_i\}_i$, $T_k$) \label{object_j}

    found, $P$ = \FindSpace($\{C_i\}_{i\neq j}\cup\{T_k\}$) \label{start:move_j}

    \If{found}{

      \KwRet ($C_j$, P) \label{end:move_j}

    }

    \KwRet ($C_k$, $C_k$) \label{move_k}

  }

  }
\end{algorithm}

As opposed to games where a complete game must be played before having access to the outcome (win or lose), the reward in our problem is defined in every state. Therefore, we found that using a MCTS simulation stage is not necessary.

The number of MCTS iterations to build the tree is typically a hyper-parameter. In order to have a generic method that works regardless of the number of objects, we adopt the following strategy: we run MCTS until a maximum (large) number of iterations is reached or until a solution is found.
We indeed noticed that the first solution provided by MCTS is already sufficiently good compared to the next ones when letting the algorithm run for longer.

The presented approach only considers tabletop rearrangement planning with
overhand grasps. The main limitation is that we assume the motion planning
algorithm can successfully plan all the pick-and-place motions computed in
Algo.~\ref{algo:action_param}. This assumption does not hold for more complex
environment where some objects are not reachable at any time (e.g. moving
objects in a constrained space such as inside a refrigerator). In this case, the function $\textsc{IS\_MOVE\_VALID}$ can be adapted to check whether the movement can be executed on the robot. Note that we consider simple collision models in $\textsc{FIND\_POSITION}$ but meshes of the objects and environment could be used if available.

\section{Experiments}

\label{sec:experiments}

We start by evaluating planning (section~\ref{sec:exmcts}) and visual scene state estimation (section~\ref{sec:exvis}) separately, demonstrating that: (i) our MCTS task planner scales well with the number of objects and is efficient enough so that it can be used online (i.e. able to recompute a plan after each movement); (ii) the visual system detects and localizes the objects with an accuracy
sufficient for grasping. %
Finally, in section~\ref{sec:exrobot} we evaluate our full pipeline in challenging setups and demonstrate that it can efficiently perform the task and can recover from errors and perturbations as also shown in the supplementary video. %

\subsection{MCTS planning}
\label{sec:exmcts}
\noindent \textbf{Experimental setup.} To evaluate planning capabilities and efficiency we {randomly} sample 3700 initial and target configurations for 1 up to 37 objects in the workspace. It is difficult to go beyond 37 objects as it becomes hard to find valid configurations for more due to the workspace constraints.

\begin{figure}
  \centering
  \includegraphics[width=1.03\columnwidth]{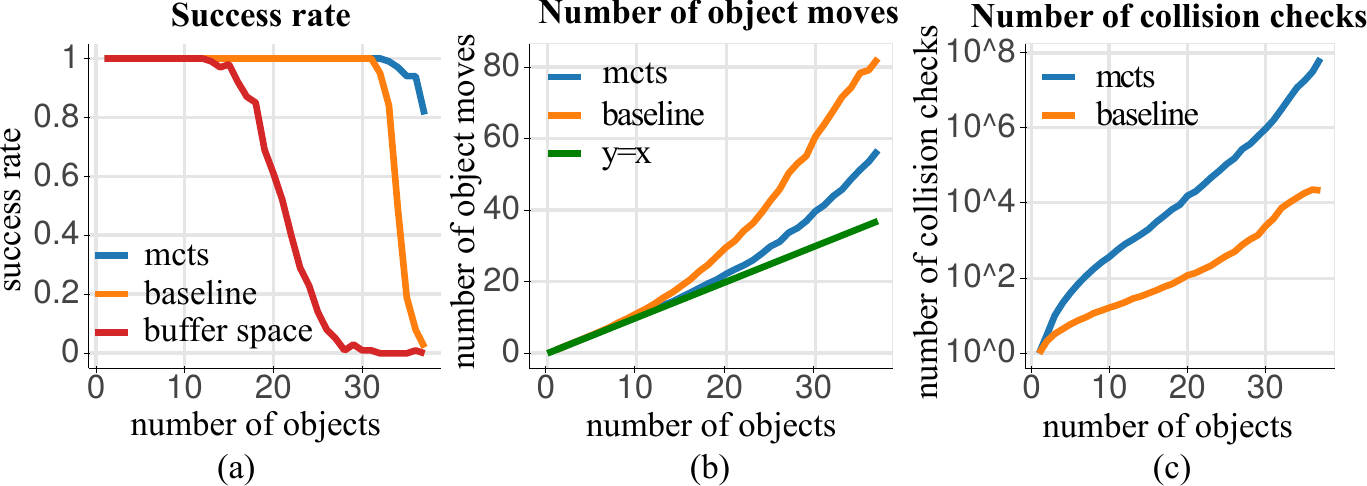} 
  \caption{\small Comparison of the proposed MCTS planning approach against the strong baseline heuristic. MCTS is able to solve more complex scenarios (with more objects) in a significantly lower number of steps. MCTS does not require free space to be available or that the problems are monotone.}
  \label{fig:comparisons}
  \vspace{-1.5em}
\end{figure}

\noindent \textbf{Planning performance.} We first want to demonstrate the interest of MCTS-based exploration compared to a simpler solution in term of efficiency and performances. As a baseline, we propose a fast heuristic search method, which simply iterates over all actions once in a random order, trying to complete the rearrangement using the same action space as our MCTS approach, until completion or a time limit is reached. Instead of moving only the closest object that is obstructing a target position $T_k$, we move all the objects that overlap with $T_k$ to their target positions or to a free position inside the workspace that do not overlap with $T_k$. Our MCTS approach is compared with this strong baseline heuristic in Fig.~\ref{fig:comparisons}. Unless specified otherwise, we use $c=1$ for MCTS and set the maximum number of MCTS iterations to 100000.

As shown in Fig.~\ref{fig:comparisons}(a), the baseline is able to solve complex instances but its success rate starts dropping after 33 objects whereas MCTS is able to find plans for 37 objects with 80\% success.
More importantly, as shown in Fig.~\ref{fig:comparisons}(b), the number of object movements in the plans found by MCTS is significantly lower.
For example, rearranging 30 objects takes only 40 object moves with MCTS compared to 60 with the baseline. 
This difference corresponds to more than 4 minutes of real operation in our robotics setup. 
The baseline and MCTS have the same action parametrization but MCTS produces higher quality plans because it is able to take into consideration the future effects of picking-up an object and placing it at a specific location.
On a laptop with a single CPU core, MCTS finds plans for 25 objects in 60\,ms. This high efficiency allows to replan after each pick-and-place action and perform rearrangement in a closed loop.

\begin{figure}
  \centering
  \includegraphics[width=1.03\columnwidth]{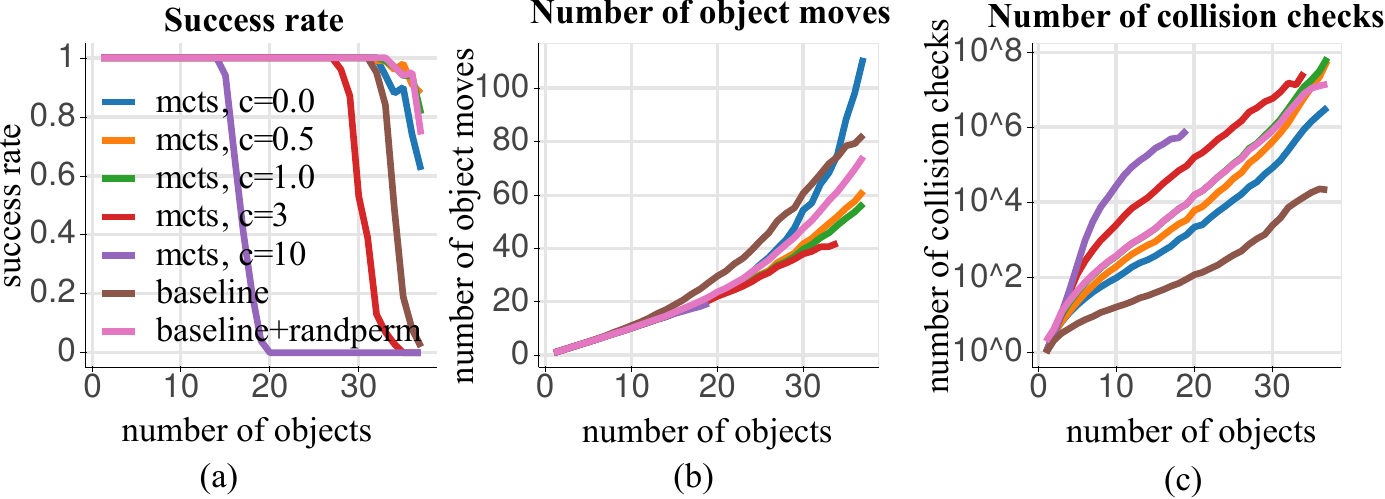} 
  \caption{\small Exploration-exploitation tradeoff in MCTS. MCTS performs better than a random baseline heuristic search. Balancing the exploration term of UCB with the parameter $c$ is crucial for finding good solutions while limiting the number of collision checks.}

  \label{fig:rl-tradeoff}
  \vspace{-1.5em}
\end{figure}

\noindent \textbf{Exploration-exploitation trade-off.} We now want to demonstrate that the benefits of our planning method are due to the quality of the exploration/exploitation trade-off in MCTS. An important metric is the total number of collision checks that the method requires for finding a plan. The collision check (checking whether an object can be placed to a certain location) is indeed one of the most costly operation when planning.
Fig.~\ref{fig:comparisons}(c) shows that MCTS uses more collision checks compared to the baseline because MCTS explores many possible optimistic action sequences while the baseline is only able to find {a} solution and does not optimize any objective. We propose another method that we refer to as \textit{baseline+randperm} which solves the problem with the baseline augmented with a random search over action sequences: the baseline is run with different random object orders until a number of collision checks similar to MCTS with $c=1$ is reached and we keep the solution which has the smallest number of object moves. As can be see in Fig.~\ref{fig:rl-tradeoff}, baseline+randperm has a higher success rate and produces higher quality plans with lower number of object moves compared to the baseline (Fig.~\ref{fig:rl-tradeoff}(b)). However, MCTS with $c=1$, still produces higher quality plans given the same amount of collision checks. 
The reason is that baseline+randperm only relies on a random exploration of action sequences while MCTS allows to balance the exploration of new actions with the exploitation of promising already sampled partial sequences through the exploration term of UCB (equation~\ref{eq:uct_uniform}). In Fig.~\ref{fig:rl-tradeoff}, we also study the impact of the exploration parameter $c$. MCTS with no exploration ($c=0$) finds plans using fewer collision checks compared to $c>0$ but the plans have high number of object moves. Increasing $c$ leads to higher quality plans while also increasing the number of collision checks. Setting $c$ too high also decreases the success rate (c=3, c=10 in Fig~\ref{fig:rl-tradeoff}(a)) because too many nodes are added to the tree and the limit on the number of MCTS iterations is reached before finding a solution.

\begin{figure}
  \centering
  \includegraphics[width=1.03\columnwidth]{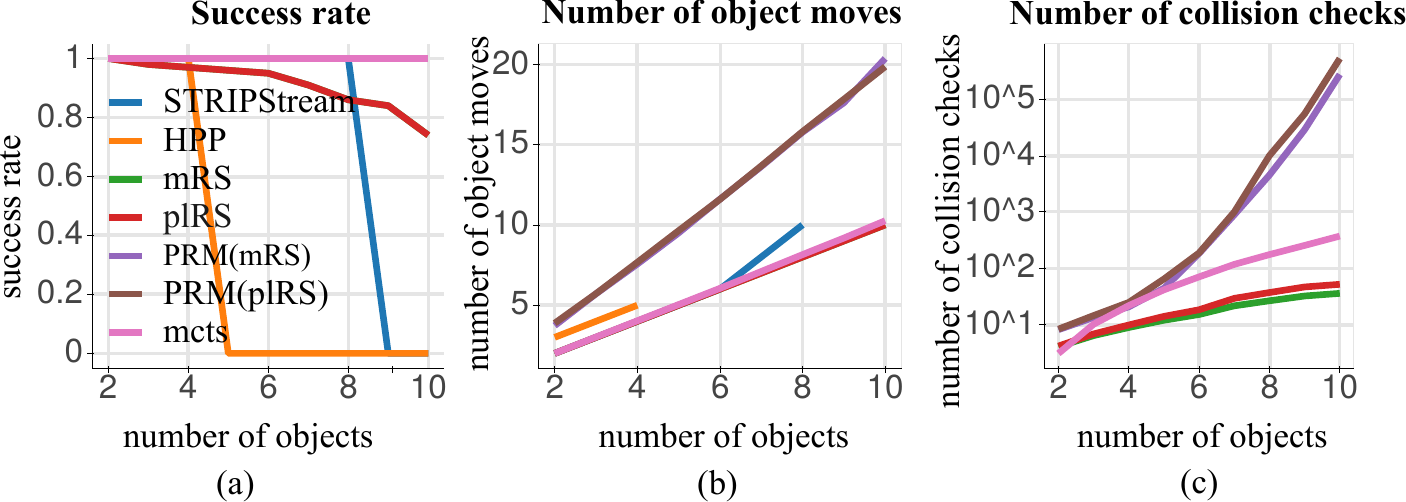} 
  \caption{\small Comparison of our MCTS planning approach with several other state-of-the-art methods. MCTS performs better than other methods applied to the rearrangement planning problem. MCTS finds high quality plans (b) using few collisions checks (c) with 100\% success rate for up to 10 objects.}
  \vspace{-1em}
  \label{fig:baselines}
\end{figure}

\noindent \textbf{Generality of our set-up.} Previous work for finding high-quality solutions to rearrangement problems has been limited to either monotone instances \cite{krontiris2016efficiently} or instances where buffer space is available \cite{Han2017-sa}. 
The red curve in Fig.~\ref{fig:comparisons}(a) clearly shows that in our set-up the number of problems where some buffer space is available for at least one object quickly decreases with the number of objects in the workspace. %
In other words, the red curve is an upper bound on the success rate of
\cite{Han2017-sa}, which requires available buffer space.
In order to evaluate the performance of our approach on monotone problems, we generated the same number of problems but the target configuration was designed by moving object from the initial configuration one by one, in a random order into free space. This ensures that the instances are monotone and can be solved by moving each object once. Our MCTS-based approach was able to solve 100\% of these instances optimally in $N$ steps. 
Our method can therefore solve the problems considered in~\cite{krontiris2016efficiently} while also being able to handle significantly more challenging non-monotonic instances, where objects need to be moved multiple times.

\noindent \textbf{Comparisons with other methods.}  To demonstrate that other planning methods do not scale well when used in a challenging scenario similar to ours, we compared our planner with three other methods of the state of the art: STRIPStream \cite{Garrett2018-te}, the Humanoid Path-Planner~\cite{Mirabel2016-zu}, mRS and plRS \cite{krontiris2015dealing}. 
Results are presented in Fig.~\ref{fig:baselines} for 900 random rearrangement
instances, we limit the experiments to problems with up to 10 objects as evaluating these
methods for more complex problems is difficult given a reasonable amount of
time (few hours). HPP~\cite{Mirabel2016-zu} is the slowest method and could not handle more than 4 objects, taking more than 45 minutes of computation for solving the task with 4 objects. HPP fails to scale because it attempts to solve the combined task and motion planning problem at once using RRT without explicit task/motion planning hierarchy thus computing many unnecessary robot/environment collision checks. The other methods adopt a task/motion planning hierarchy and we compare results for the task planners only. 
The state-of-the-art task and motion planner for general problems, STRIPSStream~\cite{Garrett2018-te}, is able to solve problems with up to 8 objects in few seconds but do not scale (Fig.~\ref{fig:baselines}(a)) when target locations are specified for all objects in the scene as it is the case for rearrangement planning problems. The success rate of specialized rearrangement methods, mRS and plRS, drops when increasing number of objects because these methods cannot handle situations where objects are permuted, i.e. placing an object at its target position requires moving another objects first thus requiring longer term planning capability. When used in combination with a PRM, more problems can be addressed but the main drawback is that these methods are slow as they perform a high number of collision checks (Fig.~\ref{fig:baselines}(c)). Overall, the main limitation of STRIPSTream,  PRM(mRS) and PRM(plRS) comes from the fact that the graph of possible states is sampled randomly whereas MCTS will prioritize the most optimistic branches (exploitation). MCTS and it's tree structure also allows to build the action sequence progressively (moving one object at once) compared to PRM-based approaches that sample entire arrangements and then try to solve them.

\subsection{Visual scene state estimation}
\label{sec:exvis}

\begin{figure}
  \centering
  \includegraphics[width=1.03\columnwidth]{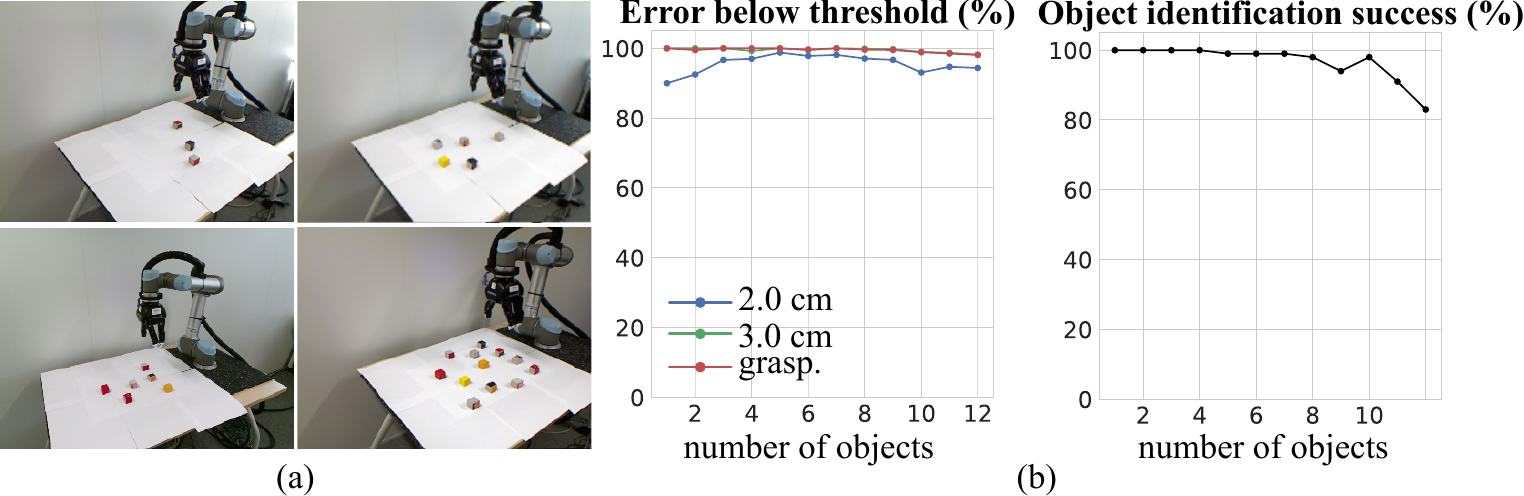} 
  \caption{\small (a) Example images from the dataset that we use to evaluate the accuracy of object position estimation as
    a function of number of objects. (b) Evaluation of our visual system for a variable number of
    objects. We report the localization accuracy (left) and the percentage of images where all objects are
    correctly identified (right). Please see the supplementary video for
      examples where our system is used in more challenging visual conditions.}
  \vspace{-1.8em}
  \label{fig:vision_results}
  \label{fig:test_n_objects}
  \label{fig:plot_n_objects}
\end{figure}

\noindent \textbf{Experimental setup.}
To evaluate our approach, we created a dataset of real images with known object configurations. We used 1 to 12 small 3.5 cm cubes, 50 configurations for each number of cubes, and captured images using two cameras for each configuration, leading to a total of 1200 images depicting 600 different configurations. 
Example images from this evaluation set are shown in
Fig.~\ref{fig:test_n_objects}(a).

\noindent \textbf{Single object accuracy.}
When a single object is present in the image, the mean-shift algorithm always succeeds and the precision of our object position prediction is \mbox{$1.1\,\pm\,0.6$\,cm}. This is comparable to the results reported in \cite{Tobin2017DomainRF} for positioning of a known object with respect to a known table without occlusions and distractors, $1.3\,\pm\,0.6$\,cm, and to results reported in~\cite{Loing2018} for the coarse alignment of a single object with respect to a robot, \mbox{$1.7\,\pm\,3.4$\,cm}. The strength of our method, however, is that this accuracy remains constant for up to 10 previously unknown objects, a situation that neither \cite{Tobin2017DomainRF} nor \cite{Loing2018} can deal with.

\noindent \textbf{Accuracy for multiple objects.}
In Fig.~\ref{fig:plot_n_objects}(b), we report the performance of the object localization and object identification modules as a function of the number of objects. For localization, we report the percentage of objects localized with errors below 2\,cm and 3\,cm respectively. For 10 objects, the accuracy is $1.1\,\pm\,0.6$\,cm. The 3\,cm accuracy approximately corresponds to the success of grasping, that we evaluate using a simple geometric model of the gripper. 
Note that grasping success rates are close to 100\% for up to 10 objects. As the number of objects increases, the object identification accuracy decreases slightly because the objects start to occlude one each other in the image. 
This situation is very challenging when the objects are unknown because two objects that are too close can be perceived as a single larger object. Note that we are not aware of another method that would be able to directly predict the workspace position of multiple unseen objects with unknown dimensions using only (uncalibrated) RGB images as input.

\noindent \textbf{Discussion.}
Our experiments demonstrate that our visual predictor is able to scale well up
to 10 objects with a constant precision that is sufficient for grasping.
We have also observed that our method is able to generalize to objects with shapes not seen during training, such as cups or plastic toys. While we apply our visual predictor to visually guided rearrangement planning, it could be easily extended to other contexts using additional task-specific synthetic training data.
Accuracy could be further improved using a refinement similar to \cite{Loing2018}. Our approach is limited to 2D predictions for table-top rearrangement planning. Predicting 6DoF pose of unseen  objects precise enough for robotic manipulation remains an open problem.

\subsection{Real robot experiments using full pipeline}
\label{sec:exrobot}

We evaluated our full pipeline, performing both online visual scene estimation
and rearrangement planning by performing 20 rearrangement tasks, each of them composed with 10
objects. In each case, the target configuration was described by an image of a
configuration captured from a different viewpoint,
  with a different table texture and a different type of camera. 
  Despite the very challenging nature of the task, our system succeeded in correctly solving $17/20$ of the experiments. In case of success, our system used on average 12.2 steps. The three failures were due to errors in the visual recognition system (incorrectly estimated number of objects, mismatch of source and target objects).   
Interestingly, the successful cases were not always perfect runs, in the sense that
the re-arrangement strategy was not optimal or that the visual estimation confused two objects at one step of the matching process. However, our
system was able to recover robustly from these failures because it is applied in a closed-loop fashion, where then plan is recomputed at each object move.

{\bf The supplementary video shows additional experiments} including objects other than cubes, different backgrounds, a moving hand-held camera and
 external perturbations, where an object is moved during the rearrangement. These results demonstrate the robustness of our system. 
 To the best of our knowledge, rearranging a priory unknown number of unseen objects with a robotic arm while relying only on images captured by a moving hand-held camera and dealing with object perturbations has not been demonstrated in prior work.

\vspace{-1.2em}
\section{Conclusion}
\label{sec:conclusion}

We have introduced a robust and efficient system for online rearrangement planning, that scales to many objects and recovers from perturbations, without requiring calibrated camera or fiducial markers on objects. To our best knowledge, such a system has not been shown in previous work. At the core of our approach is the idea of applying MCTS 
to rearrangement planning, which leads to better plans, significant speed-ups and ability to address more general set-ups compared to prior work.
While in this work we focus on table-top re-arrangement, the proposed MCTS approach is general and opens-up the possibility for efficient re-arrangement planning in 3D or non-prehensile set-ups.

{\small
\bibliographystyle{IEEEtran}
\bibliography{IEEEabrv,root}
}

\onecolumn

\end{document}